%% file: 0_main.tex
\documentclass[10pt,twocolumn,letterpaper]{article}

\usepackage{iccv}
\usepackage{times}
\usepackage{epsfig}
\usepackage{graphicx}
\usepackage{amsmath}
\usepackage{amssymb}

\usepackage{algorithm}
\usepackage{algorithmic}
\usepackage{multirow}
\usepackage{caption}
\usepackage{hhline}
\usepackage{xcolor}
\usepackage{dblfloatfix}
\usepackage{subcaption}

\usepackage[pagebackref=true,breaklinks=true,letterpaper=true,colorlinks,bookmarks=false]{hyperref}

\iccvfinalcopy 

\newcommand{\sys}{SWNet}

\ificcvfinal\fi
\begin{document}

\title{\sys{}: Small-World Neural Networks and Rapid Convergence}

\author{Mojan Javaheripi\\
UC San Diego\\
\and
Bita Darvish Rouhani\\
Microsoft Research\\
\and
Farinaz Koushanfar\\
UC San Diego
}

\maketitle

\begin{abstract}
\vspace{-0.2cm}
Training large and highly accurate deep learning (DL) models is computationally costly. This cost is in great part due to the excessive number of trained parameters, which are well-known to be redundant and compressible for the execution phase. This paper proposes a novel transformation which changes the topology of the DL architecture such that it reaches an optimal cross-layer connectivity. This transformation leverages our important observation that for a set level of accuracy, convergence is fastest when network topology reaches the boundary of a Small-World Network. Small-world graphs are known to possess a specific connectivity structure that enables enhanced signal propagation among nodes. Our small-world models, called \sys{}s, provide several intriguing benefits: they facilitate data (gradient) flow within the network, enable feature-map reuse by adding long-range connections and accommodate various network architectures/datasets. Compared to densely connected networks (e.g., DenseNets), \sys{}s require a substantially fewer number of training parameters while maintaining a similar level of classification accuracy.
We evaluate our networks on various DL model architectures and image classification datasets, namely, CIFAR10, CIFAR100, and ILSVRC (ImageNet). Our experiments demonstrate an average of $\approx 2.1\times$ improvement in convergence speed to the desired accuracy.

\end{abstract}

\input{1_intro}

\input{2_related}
\input{3_prelim}
\input{5_SWN_DNN}

\input{7_experiments}

\input{8_discussion}
\input{9_conclusion}

{\small
\bibliographystyle{ieee}
\bibliography{egbib}
}

\end{document}

%% file: 1_intro.tex
\vspace{-0.5cm}
\section{Introduction}
Deep learning models are increasingly popular for various learning tasks, particularly in visual computing applications. A big advantage for DL is that it can automatically learn the relevant features by computing on a large corpus of data, thus, eliminating the need for hand-selection of features common in traditional methods. In the contemporary big data realm, visual datasets are increasingly growing in size and variety. For instance, the ILSVRC challenge dataset has 22K classes with over 14M images~\cite{russakovsky2015imagenet}. To increase inference accuracy on such challenging datasets, DL models are evolving towards higher complexity architectures. State-of-the-art models tend to reach good accuracy, but they suffer from a dramatically high training cost.

\vspace{0.5em}
As DL models grow deeper and more complex, the large number of stacked layers gives rise to a variety of problems, e.g., vanishing gradients~\cite{glorot2010understanding, bengio1994learning}, which renders the models hard to train. To facilitate convergence and enhance the gradient flow for deeper models, creation of bypass connections was recently suggested. These shortcuts connect the layers that would otherwise be disconnected in a traditional Convolutional Neural Network (CNN)~\cite{highwaynets,he2016deep,huang2017densely,xie2017aggregated}. To curtail the cost of hand-crafted DL architecture exploration, the existing literature typically realizes the shortcuts by replicating the same building block throughout the network~\cite{he2016deep,huang2017densely,xie2017aggregated}. However, such repeated pattern of blocks in these networks induces unnecessary redundancies~\cite{huang2017condensenet} that increase the computational overhead.

This paper proposes a novel methodology that transforms the topology of conventional CNNs such that they reach optimal cross-layer connectivity. This transformation is based on our observation that the pertinent connectivity pattern highly impacts training speed and convergence. To ensure computational efficiency, our architectural modification takes place \textbf{prior} to training. Thus, the incorporated connectivity measure must be independent of network gradients/loss and training data. Towards this goal, we view CNNs as graphs and revisit Small-World Networks (SWNs)~\cite{watts1998collective} from graph theory to transform CNNs into highly-connected small-world topologies. Watts-Strogatz SWNs~\cite{watts1998collective} are widely used in the analysis of complex graphs; Due to SWNs' specific connection pattern, these structures provide theoretical guarantees for considerably decreased consensus times~\cite{olfati2005ultrafast, tahbaz2007small, zanette2002dynamics}. 
\begin{figure}[h]
    \centering
    \vspace{0.5em}
    \includegraphics[width=1\columnwidth]{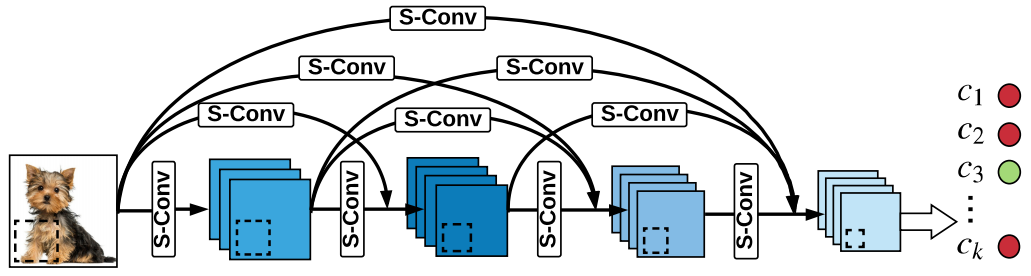}
    \caption{Schematic representation of the connections within a small-world CNN. An arbitrary neuron's output is connected to selected neurons in proceeding layers via sparse connections (convolutions) denoted by \textit{S-CONV}.}
    \label{fig:global_flow}
\end{figure}

Our network modification algorithm takes as input a conventional CNN architecture and  enforces the small-world property on its topology to generate a new network, , called \sys{}. We leverage a quantitative metric for \textit{small-worldness} and devise a customized rewiring algorithm. Our algorithm restructures the inter-layer connections in the input CNN to find a topology that balances \textit{regularity} and \textit{randomness}, which is the key characteristic of SWNs~\cite{watts1998collective}. Small-world property in CNNs translates to an architecture where all layers are interlinked via sparse connections. An example of such network is shown in Fig.~\ref{fig:global_flow}. 

\sys{}s have similar quality of prediction and number of trainable parameters as their baseline feed-forward architectures, but due to the added sparse links and the optimal SWN connectivity, they warrant better data flow. In summary, our architecture modification has three main properties: (i)~It removes non-critical connections and reduces computational implications. (ii)~It increases the degrees of freedom during training, allowing faster convergence. (iii)~It provides customized data paths in the model for better cross-layer information propagation. 

We conduct comprehensive experiments on various network architectures and showcase \sys{}s' performance on popular image classification benchmarks including CIFAR10, CIFAR100, and ImageNet. Our small-world CNNs achieve an average of 2.1-fold improvement in training iterations required to achieve comparable classification accuracy as the baseline models.
We further compare \sys{} with the state-of-the-art DenseNet model and show that with $10\times$ fewer parameters, \sys{}s demonstrate identical performance during training. 


%% file: 2_related.tex
\section{Related Work and Background}
\begin{figure}[t]
    \centering
    \includegraphics[width=0.85\columnwidth]{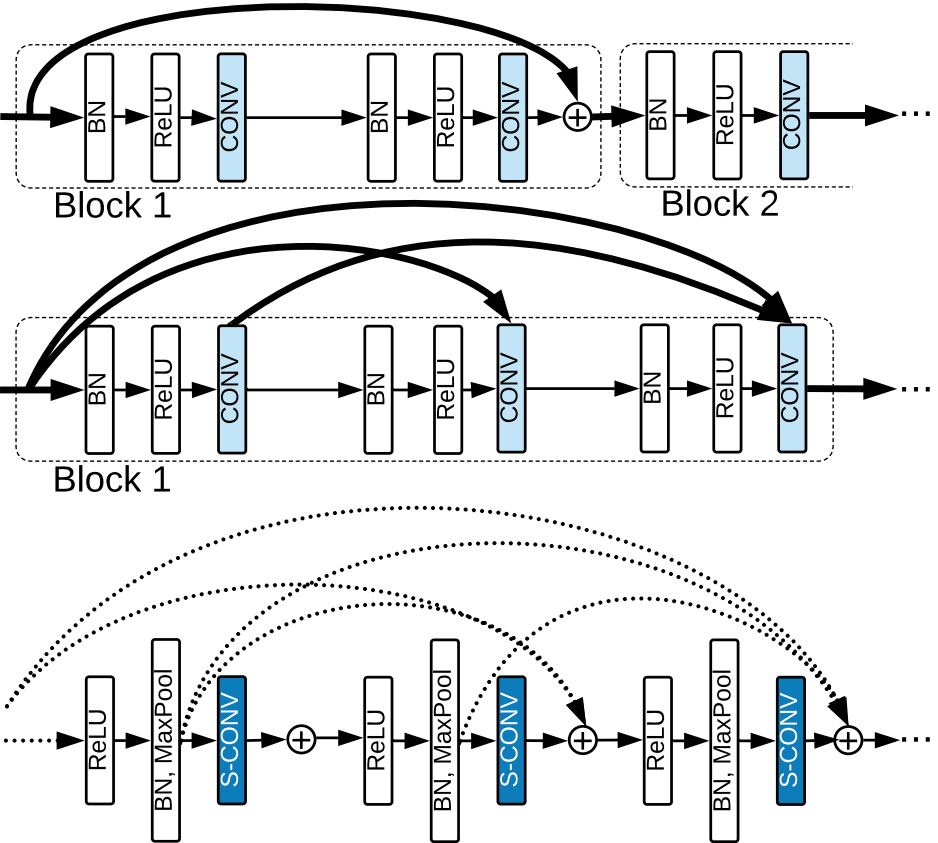}
    \vspace{-0.5em}
    \caption{Information flow within a ResNet~(top), DenseNet~(middle), and \sys{} (bottom) network. Here, \textit{CONV}, \textit{BN}, \textit{ReLU} denote a convolution kernel, batch normalization, and non-linear activation, respectively, and \sys{} customized sparse convolutions are shown as \textit{S-CONV}. Normal inter-layer connections are represented with bold lines and dotted lines are \sys{} selective inter-links.}
    \label{fig:blocks}
    \vspace{-1.5em}
\end{figure}

\subsection{Related work}
\noindent\textbf{Bypass Connections.} A substantial amount of research has focused on the addition of bypass connections to the hierarchical CNN architecture to enhance inter-layer information flow and enable feature reuse. Authors of ~\cite{highwaynets} implement the bypass connections using parametrized (gated) interlinks to enable model fine-tuning. In order to avert the burst in the number of trainable parameters caused by such gated connections, ResNets~\cite{he2016deep} use identity links (skip connections) to connect the concatenated layers. Such skip connections follow a modular structure. There exist a significant amount of redundancy in (deep) ResNets as alternative inter-layer connections may exist that render higher accuracy while having lower model complexity; as shown by~\cite{huang2016deep}, not all identity links are necessary. 

A variation of ResNets that uses wider residual blocks is introduced in ~\cite{targ2016resnet, xie2017aggregated} to further improve image classification accuracy, while the effects of such architectural modification on the convergence speed and training overhead still need a more comprehensive study. Inception networks~\cite{inception} are another example of benefiting from wider networks. Authors of ~\cite{szegedy2017inception} show that addition of residual connections to the initially proposed inception architecture drastically increases model convergence speed. This work further motivated us to study CNN convergence gains by addition of bypass connections.    

DenseNets~\cite{huang2017densely} group CNN layers in blocks with each layer connected to all its preceding layers. This is done by concatenating previous layers' feature-maps and using it as the input. Another work~\cite{huang2017condensenet} argues that such dense connectivity pattern incurs redundancies since earlier features might not be required in later layers. The authors propose to prune such redundancies to generate a more efficient architecture for CNN inference phase. However, the paper does not explore the possible effects of pruning on training.

In summary, the prior work mainly focuses on accuracy gains of long-range connections with little attention to the training overhead induced by the introduction of redundant parameters. In contrast to prior art, we select only to add long-range connections that are key contributors to model accuracy as well as convergence speed. To the best of our knowledge, \sys{} is the first work to intertwine the small-world property with CNNs and to examine the trained network in terms of convergence speed and accuracy. 

To further highlight the distinction between our work and prior art, Fig.~\ref{fig:blocks} illustrates the connection patterns in a ResNet, DenseNet, and \sys{} architecture. In contrast to these two models, \sys{} is not structured upon fixed building blocks and therefore can adapt to any given network architecture. Different from DenseNets which only accommodate fully dense connections, \sys{} leverages customized sparse convolutions. Such sparsities enable selective connectivity between pairs of layers that enhance convergence speed while ensuring a low redundancy. 

\vspace{0.1cm}
\noindent\textbf{Small-wold Network.} Perhaps the first investigation of SWNs in the context of deep learning was performed in~\cite{erkaymaz2017performance}, where the authors transform simple MLPs to SWN graphs and study the accuracy benefits for diagnosis of diabetes. \sys{} substantially differs from this work as our solution is applicable to convolutional neural networks and uses a different mathematical model and small-worldness metric.


%% file: 3_prelim.tex
\subsection{Background: Small-World Networks}\label{sec:small_world_nets}
Watts and Strogatz~\cite{watts1998collective} observed that real-world complex networks, e.g., the anatomical connections in the brain and the neural network of animals, cannot be modeled using the existing regular or random graph classes. As such, they introduced the new category of \textit{small-world networks}. Members of the small-world class have two main characteristics: 1)~They have a small average pairwise-distance between graph nodes.
2)~Nodes within the graph exhibit a relatively high (local) clustered structure.
The first property is mainly associated with random graphs while the second property is prominent in regular graph classes. 
Such networks have shown significant enhancement in signal propagation speed, consensus, synchronization, and computational capability ~\cite{strogatz2001exploring, latora2001efficient, barahona2002synchronization, kuperman2001small, zanette2002dynamics}. 

\begin{figure}[h]
    \centering
    \includegraphics[width=0.98\columnwidth]{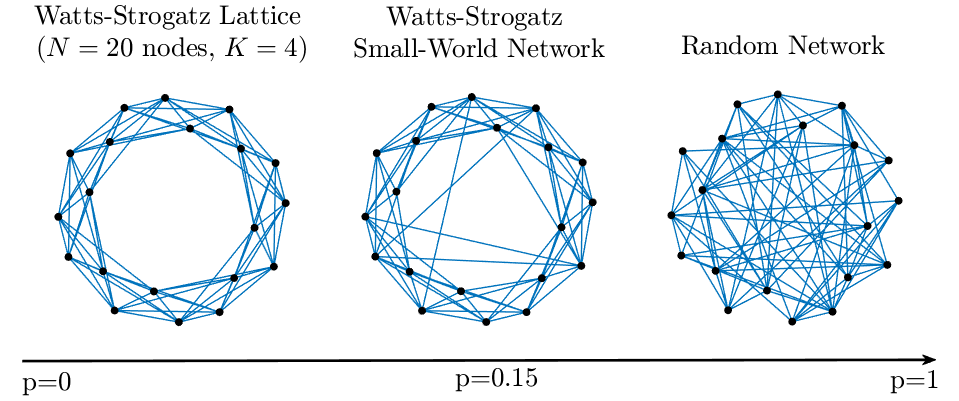}
    \caption{Transition of a regular graph to a completely random network. Intermediate values of the random rewiring probability, $p$, generate SWNs, i.e.,  clustered structures where any arbitrary node pair is connected by few edges.} 
    \label{fig:lattice}
\end{figure}

Randomness is introduced into a regular graph structure by iterative removal and addition of edges with probability, $p$, in order to construct an SWN. Fig.~\ref{fig:lattice} demonstrates the transition between a regular structure and its corresponding random graph as the rewiring probability increases from $0$ to $1$. Intermediate values of $p$ interpolate between complete regularity and randomness to generate an SWN.

%% file: 5_SWN_DNN.tex
\section{\sys{}: Small-World CNNs} \label{sec:SWN_CNNs}
\vspace{-0.2cm}
We propose to restructure the inter-layer connections in a DL model such that its topology falls into the small-world category while the total number of parameters in the network is held \textbf{constant}. 
Throughout the paper, we use the terms DL model and CNN interchangeably but emphasize that our approach is easily applicable to models without convolutions, e.g., Multi-Layer Perceptrons (MLPs).

In the following, we first elaborate on the small-world criteria and introduce methods to distinguish SWNs from other topologies (Sec.~\ref{sec:SWN_measure}). We then explain our conversion of an arbitrary CNN into its equivalent SWN (Sec.~\ref{sec:architecture_search}). Lastly, we delineate our implementation and formalize the operations performed in a \sys{} (Sec.~\ref{sec:build_swn}).

\input{5_1_metric}
\input{5_2_arch_search}
\input{5_3_methodology}

%% file: 5_1_metric.tex
\subsection{Metric for Small-Worldness}\label{sec:SWN_measure}
\vspace{-0.1cm}
To examine the small-world property for a given graph, we study two properties, namely, the \textit{characteristic path length} ($L$) and the \textit{global clustering coefficient} ($C$).
$L$ is defined as the average distance between pairs of nodes in the graph and $C$ is a measure for the density of connections between neighbors of any node in the network. 
A completely random graph lacks clustering but enjoys a small $L$. By definition, a graph is small-world if it has a similar $L$ but higher $C$ than an $Erd\ddot{o}s-Re'nyi$ ($E–R$) random graph~\cite{zhang2017random} constructed using the same number of vertices and edges. 

Let us denote the clustering coefficient and the characteristic path length of a given graph ($G$) by $C_G$ and $L_G$, respectively. In a similar fashion, we represent the corresponding characteristics of the equivalent $E–R$ random graph by $C_{rand}$, $L_{rand}$. We use a quantitative measure of the small-world property form~\cite{humphries2008network} which categorizes a network as a SWN if $S_G > 1 $ and $S_G$ is calculated using Eq.~(\ref{eq:SWN_quantitative}).
\vspace{-0.5em}
\begin{equation}\label{eq:SWN_quantitative}
    S_G=\frac{\gamma_G}{\lambda_G},~~~~\gamma_G=\frac{C_G}{C_{rand}},~~~~\lambda_G=\frac{L_G}{L_{rand}}
\end{equation}

%% file: 5_2_arch_search.tex
\subsection{Small-world Architecture Acquisition}\label{sec:architecture_search}
\subsubsection{Graph Generation} 
In order to modify a given CNN architecture and generate the equivalent SWN, we first model all connections within the network as a graph representation. In this context, a connection is defined as a linear operation performed between an input element and a trainable weight (network parameter) found in \textit{Convolution} ($Conv$) and \textit{Fully-Connected} ($FC$) layers. For $Conv$ layers, each feature-map channel is represented by a node and each edge represents a $k\times k$ kernel. For $FC$ layers each neuron is assigned a separate node and the edges correspond to weight matrix elements.


\vspace{-0.2cm}
\subsubsection{Architecture Search} 
\vspace{-0.1cm}
After generating the graph pertinent to the input CNN architecture, we proceed to find the equivalent SWN. To perform this task, the initial graph is randomly rewired with different probabilities, $p \in[0,1]$, similar to Fig.~\ref{fig:lattice}. For each rewired graph, we compute the characteristic path length  $L$ and clustering coefficient $C$ and use the captured pattern for each criterion to detect the small-world topology using the small-worldness measure defined in Sec.~\ref{sec:SWN_measure}.

\begin{figure}[ht!]
    \centering
    \includegraphics[width=0.6\columnwidth]{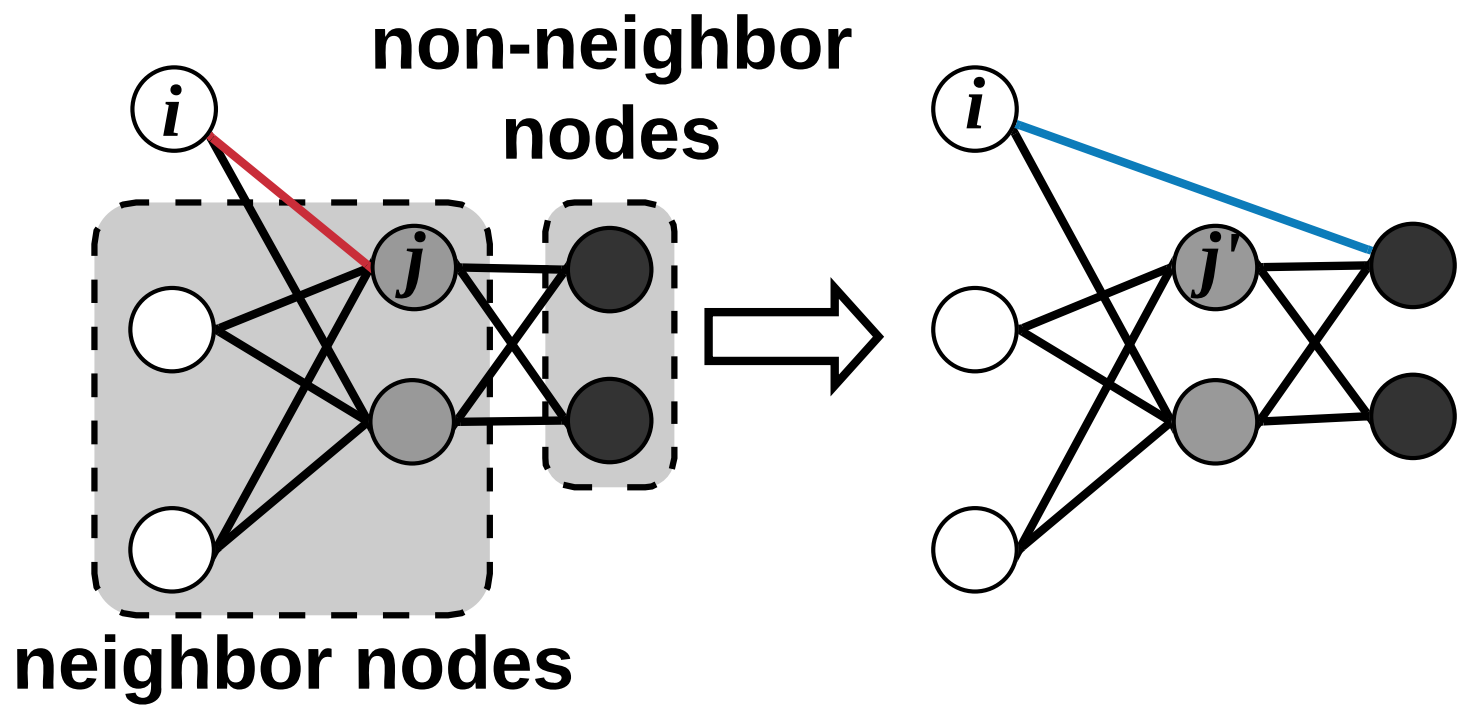}
    \vspace{-0.1cm}
    \caption{Our proposed rewiring algorithm replaces edges to the subsequent layer (red) with long-range edges (blue).}
    \label{fig:rewiring}
\end{figure}

\noindent\textbf{Rewiring Policy.} Let us denote an edge with $e(v_i, v_j)$ where $v_i$ and $v_j$ are the start and end nodes. To perform random rewiring with probability $p$, we visit all edges in the graph once. Each edge is rewired with probability $p$ or kept the same with probability $1-p$. If the edge is to be rewired, a new second node $v_{j'}$ is randomly sampled from the set of nodes that are non-neighbor to the edge's start node, $v_i$.  
This second node is selected such that no self-loops or repeated links exist in the rewired graph. 
Once the destination node is chosen, the initial edge, $e(v_i, v_j)$ is removed and replaced by $e(v_i, v_{j'})$. Fig.~\ref{fig:rewiring} demonstrates our rewiring mechanism.
Note that our rewiring methodology does not alter the number of connections in the CNN. As a result, the total number of trainable parameters in the SWN model equals that of the original network. 

\vspace{-0.1cm}

\begin{figure}[h]
    \centering
    \includegraphics[width=0.9\columnwidth]{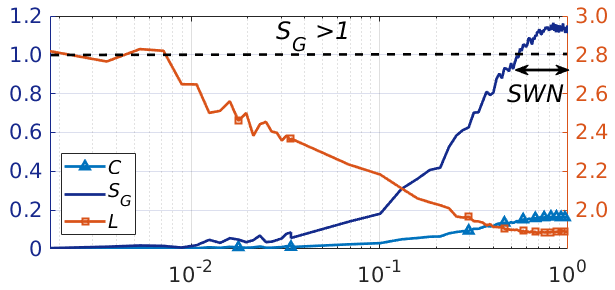}
    \vspace{-0.3cm}
    \caption{Clustering coefficient ($C$), small-world property ($S_{G}$), and path length ($L$) versus rewiring probability. The region where the graph transforms into a small-world network is shown with the double-headed arrow.}
    \label{fig:C_L_plots}
\end{figure}

\vspace{-0.2cm}
\noindent\textbf{Network Profiling.} Using the aforementioned rewiring policy, we generate various graphs by sweeping the rewiring probability in the [0,1] interval. Fig.~\ref{fig:C_L_plots} demonstrates the correlation between $C$ and $L$ as the rewiring probability is changed for a 14-layer CNN model. For conventional CNNs, the clustering coefficient is zero and the characteristic path length can be quite large specifically for very deep networks (leftmost points on Fig.~\ref{fig:C_L_plots}).
As such, CNNs are far from networks with the small-world property. Random rewiring replaces short-range connections to immediately subsequent layers with longer-range connections. Consequently, $L$ is reduced while $C$ increases as the network shifts towards the small-world equivalent. We select the topology with the \textbf{maximum} value of small-world property, $S_G$, as the \sys{}. As a direct result of such architectural modification, the new network enjoys enhanced connectivity in the corresponding CNN which results in better gradient propagation and training speedup.

\begin{figure}[h]
    \centering
    \includegraphics[width=0.7\columnwidth]{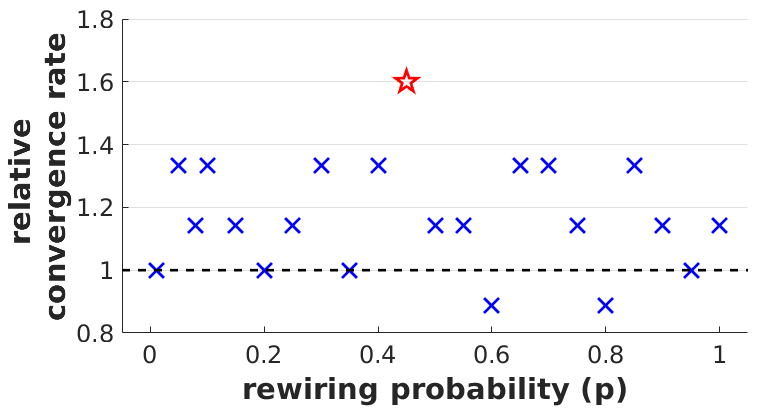}
    \caption{Convergence speed of a 5-layer CNN and its randomly rewired counterparts. All values are normalized by baseline convergence rate. SWN is shown with a red star.
    }
    \label{fig:mnist_swipe}
\end{figure}

To demonstrate the efficiency of the SWN versus other configurations generated during the probability sweep, we train several rewired networks on the MNIST dataset~\cite{lecun1998mnist}, each of which is constructed from a 5-layer CNN. Fig.~\ref{fig:mnist_swipe} demonstrates the convergence speed of these various architectures versus rewiring probability used to generate them from the baseline model. 
Due to the addition of long-range connections, almost all models show convergence improvements over the baseline. However, the perfect balance between node clustering and average path length is achieved for the SWN. This, in turn, renders the fastest convergence.

%% file: 5_3_methodology.tex
\vspace{-0.3cm}
\subsection{\sys{} Methodology}\label{sec:build_swn}

\noindent\textbf{CNN Formulation.} 
Conventional CNNs are comprised of subsequent layers where each layer, $l$, in the network performs a combination of linear and nonlinear operations on its input, $x_l$, to generate the corresponding output, $y_l$. 
We denote core linear operations ($Conv$ and $FC$) in a CNN by $W_l(\cdot)$ with the subscript representing the layer index. 
Other operations can take the form of Batch Normalization~($BN$)~\cite{ioffe2015batch}, Rectified Linear Unit ($ReLU$)~\cite{glorot2011deep}, and Pooling~\cite{lecun1998gradient}. For each linear layer, we bundle one or more of such operations together and show them as one composite function, $C_l(.)$. 
For an arbitrary layer $l$ in a conventional CNN, the output is formalized as:
\vspace{-0.2em}
\begin{equation}\label{eq:cnn_layer}
    y_l = C_l(W_l(x_l))
\end{equation}
Note that the cascaded nature of CNNs implies that the generated output from one layer serves as the input to the immediately succeeding layer: $x_{l+1}=y_l$.

\begin{figure}[h]
    \centering
    \includegraphics[width=0.95\columnwidth]{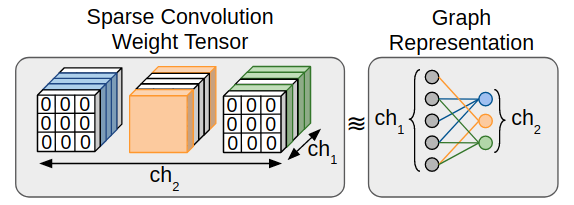}
    \vspace{-0.5em}
    \caption{Coarse-grained sparse convolution between a layer with $ch_1=5$ output channels and a layer with $ch_2=3$ output channels. Left: Sparse convolution weights. For each removed connection from the graph, the corresponding filter in the sparse convolution weight is masked to zero. Right: Equivalent graph with nodes representing channels.}
    \label{fig:sparse_conv}
\end{figure}

\vspace{0.2cm}
\noindent\textbf{Sparse Connections in \sys{}s.} One major difference between \sys{}s and conventional CNNs is that \sys{} layers can be interconnected regardless of their position in the network hierarchy. More specifically, the output of each layer of a \sys{} is connected to all its succeeding layers via sparse weight tensors. These connections are implemented via \textit{convolution} kernels with coarse-grained sparsity patterns. Fig.~\ref{fig:sparse_conv} shows the convolution filters of an example sparse connection from a layer with 5 output channels to a layer with 3 output channels and its small-world graph representation.  
Let us denote sparse connections from layer $l1$ to layer $l2$ by $W_{l_1l_2}^s(.)$. The output of the $l$-th layer in \sys{} can then be calculated as:
\begin{equation}\label{eq:sys_layers}
    y_l = C_l(W_l^s(x_l) + \sum_{l_1<l-1}{W_{l_1~l}^s(y_{l_1})})
\end{equation}
Comparing the above formulation with Eq.~\ref{eq:cnn_layer}, we highlight the extra summation term that accounts for the inter-layer connections. Note that in Eq.~\ref{eq:sys_layers}, both $W_l^s$ and $W_{l_1~l}^s$ are sparse tensors. 
The inter-layer connectivity in \sys{} enables enhanced data flow, both in inference and training stages, while the sparse connections mitigate unnecessary parameter utilization. In contrast to the previously proposed feature concatenation methodology~\cite{huang2017densely}, we perform summation over the feature-maps. By means of this approach, we mitigate the appearance of extremely high dimensional kernels that result from channel-wise feature-map concatenation. Furthermore, the summation of feature-maps enables \sys{} to be applicable to all network architectures with various layer configurations.


\vspace{0.1cm}
\noindent\textbf{Composite Non-linear Operation.}
In contrast to DenseNets~\cite{huang2017densely} and ResNets~\cite{he2016deep} where several linear layers are concatenated before pooling is performed, \sys{}s support pooling immediately after each $Conv$ layer as seen in conventional CNN architectures. We experiment with various configurations of the widely-used non-linear operations, i.e., \textit{BN}, \textit{ReLU}, \textit{Maxpool} to investigate the effect of ordering on network convergence. Our experiments demonstrate that \sys{} convergence is enhanced when the composite non-linear function, $C_l$ is implemented as a \textit{ReLU}, followed by \textit{Maxpool}ing, and \textit{BN} as shown in Fig.~\ref{fig:blocks}.


%% file: 7_experiments.tex
\begin{table*}[t]
\centering
\caption{Benchmarked CNNs for evaluating \sys{} effectiveness. $Conv$ layers are represented as $\langle kernel~size\rangle$Conv. $FC$ layers are denoted by $\langle output~elements\rangle$FC and $BN$ and $ReLU$ are not shown for brevity.
}
\label{tab:architectures}
\resizebox{1\textwidth}{!}{
\begin{tabular}{l|c|c|c|c|c|c|c|c|c|c|c}
\hline
          & \textbf{Convolution}  & \textbf{MaxPool}  & \textbf{Convolution}  & \textbf{MaxPool}     
          & \textbf{Convolution}  & \textbf{MaxPool}  & \textbf{Convolution}  & \textbf{MaxPool}     
          & \textbf{Convolution}  & \textbf{MaxPool}  & \textbf{Classifier} \\ \hline
\textbf{\begin{tabular}[c]{@{}c@{}}ConvNet-C~$^*$\cite{simonyan2014very} \\ (C10, C100)\end{tabular}} & \begin{tabular}[c]{@{}c@{}} [$3\times 3$ Conv] \\ $\times 2$\end{tabular}                                               & \begin{tabular}[c]{@{}c@{}}$2 \times 2$ \\ stride 2\end{tabular} & \begin{tabular}[c]{@{}c@{}}[$3\times 3$ Conv] \\ $\times 2$\end{tabular} & \begin{tabular}[c]{@{}c@{}}$2 \times 2$ \\ stride 2\end{tabular} & \begin{tabular}[c]{@{}c@{}}[$3\times 3$ Conv] \\ $\times 3$\end{tabular} & \begin{tabular}[c]{@{}c@{}}$2 \times 2$ \\ stride 2\end{tabular} & \begin{tabular}[c]{@{}c@{}}[$3\times 3$ Conv] \\ $\times 3$\end{tabular} & \begin{tabular}[c]{@{}c@{}}$2 \times 2$ \\ stride 2\end{tabular} & \begin{tabular}[c]{@{}c@{}}[$3\times 3$ Conv] \\ $\times 3$\end{tabular} & \begin{tabular}[c]{@{}c@{}}$2 \times 2$ \\ stride 2\end{tabular} & \begin{tabular}[c]{@{}c@{}}512FC\\ 10FC, softmax\end{tabular}                            \\ \hline

\textbf{\begin{tabular}[c]{@{}c@{}}AlexNet~\cite{krizhevsky2012imagenet} \\ (ImageNet)\end{tabular}}   & \begin{tabular}[c]{@{}c@{}}$11\times 11$ Conv\\ (stride 4)\end{tabular} & \begin{tabular}[c]{@{}c@{}}$2 \times 2$ \\ stride 2\end{tabular} & $5\times 5$ Conv          & \begin{tabular}[c]{@{}c@{}}$2 \times 2$ \\ stride 2\end{tabular} & $3\times 3$ Conv          & - & $3\times 3$ Conv      & - & $3\times 3$ Conv          & \begin{tabular}[c]{@{}c@{}}$2 \times 2$ \\ stride 2\end{tabular} & \begin{tabular}[c]{@{}c@{}}4096FC\\ 4096FC\\ 1000FC, softmax\end{tabular} \\ \hline

\textbf{\begin{tabular}[c]{@{}c@{}}ResNet-18~\cite{he2016deep} \\ (ImageNet)\end{tabular}}   & \begin{tabular}[c]{@{}c@{}}$7\times 7$ Conv\\ (stride 2)\end{tabular} & \begin{tabular}[c]{@{}c@{}}$3 \times 3$ \\ stride 2\end{tabular} & \begin{tabular}[c]{@{}c@{}}[$3\times 3$ Conv] \\ $\times 2$\end{tabular}  & - & \begin{tabular}[c]{@{}c@{}}[$3\times 3$ Conv] \\ $\times 2$\end{tabular}  & - & \begin{tabular}[c]{@{}c@{}}[$3\times 3$ Conv] \\ $\times 2$\end{tabular}  & - & \begin{tabular}[c]{@{}c@{}}[$3\times 3$ Conv] \\ $\times 2$\end{tabular}  & \begin{tabular}[c]{@{}c@{}}$7 \times 7$ \\ average pool \end{tabular} & 1000FC, softmax \\ \hline
\end{tabular}}
\vspace{0.5em}
\scriptsize{$^*$ We modify the ConvNet-C fully-connected layers form~\cite{simonyan2014very} to comply with the CIFAR datasets.
}

\resizebox{0.95\textwidth}{!}{
\begin{tabular}{c|c|c|c|c|c|c|c}
\hline
\multicolumn{1}{l|}{}                                                        & \multicolumn{1}{l|}{\textbf{Convolution}} & \multicolumn{1}{l|}{\textbf{Dense Block (1)}} & \multicolumn{1}{l|}{\textbf{Transition Block}} & \multicolumn{1}{l|}{\textbf{Dense Block (2)}} & \multicolumn{1}{l|}{\textbf{Transition Block}} & \multicolumn{1}{l|}{\textbf{Dense Block (3)}} & \multicolumn{1}{l}{\textbf{Classifier}} \\ \hline
\multirow{2}{*}{\begin{tabular}[c]{@{}c@{}}\textbf{DenseNet-40}~\cite{huang2017densely}\\ (\textbf{C10})\end{tabular}} & \multirow{2}{*}{$3\times 3$ Conv}        & \multirow{2}{*}{{[}$3\times 3$ Conv{]}$\times 12$}   & $1\times 1$ Conv                              & \multirow{2}{*}{{[}$3\times 3$ Conv{]}$\times 12$}   & $1\times 1$ Conv & \multirow{2}{*}{{[}$3\times 3$ Conv{]}$\times 12$}   &  $8\times 8$ average pool               \\ \cline{4-4} \cline{6-6} \cline{8-8} 
&  &   & $2\times 2$ average pool                      &                & $2\times 2$ average pool                      &                       & 10FC, softmax                  \\ \hline
\end{tabular}
}

\scriptsize{$^*$ Conv denotes a $BN$, followed by a $ReLU$ and a convolution layer.
}
\end{table*}

\vspace{-0.2cm}
\section{Experiments}\label{sec:experiments}
We conduct proof-of-concept experiments on different network architectures and image classification benchmarks to empirically demonstrate the enhanced convergence speed of \sys{}s compared to the baseline (conventional) counterparts. Our implementations are available in popular neural network development APIs, Keras~\cite{chollet2015keras} and PyTorch~\cite{paszke2017automatic}.

\subsection{Datasets}\label{sec:dataset}
\noindent\textbf{CIFAR.} We carry out our experiments on the two available CIFAR~\cite{cifar} datasets. CIFAR10~(C10) and CIFAR100~(C100) benchmarks consist of colored images with dimensionality $32\times 32$ that are categorized in 10 and 100 classes, respectively. Each dataset contains 50,000 samples for training and 10,000 samples for testing. We use standard data augmentation routines popular in prior work~\cite{he2016deep,huang2016deep}. The samples are normalized using per-channel mean and standard deviation. At training time, random horizontal mirroring, shifting, and slight rotation are also applied.

\vspace{0.1cm}
\noindent\textbf{ImageNet.} The ISLVRC-2012 dataset, widely known as the ImageNet~\cite{deng2009imagenet}, consists of 1000 different classes of colored images with 1.2 million samples for training and 50,000 samples for validation. We use the augmentation scheme proposed in~\cite{simonyan2014very,he2016identity} to preprocess input samples. During training, we resize the images by randomly sampling the shorter edge from [256, 480].
A $224\times224$ crop is then randomly sampled from the image. We also perform per-channel normalization as well as horizontal mirroring~\cite{krizhevsky2012imagenet}.


\subsection{Benchmarked Architectures}\label{sec:architecture}
Tab.~\ref{tab:architectures} encloses our baseline CNN architectures. \sys{}s maintain the same feed-forward architecture as the baseline networks and are constructed by 1)~replacing the original $Conv$ layers with sparse convolutions and 2)~implementing additional sparse convolutions between non-consecutive layers. 
To match the dimensionality of inter-layer connected feature-maps, we tune the stride in the long-range sparse connections and use zero-padding where necessary\footnote{~We make sure that the stride is smaller than convolution window size}. This approach enables us to control the dimensionality of the produced feature-maps as well as tune the impact of added long-range connections.

\input{7_1_CIFAR}
\input{7_2_imagenet}

%% file: 7_1_CIFAR.tex
\begin{figure*}[b!]
    \centering
    \vspace{1.25em}
    \includegraphics[width=0.85\textwidth]{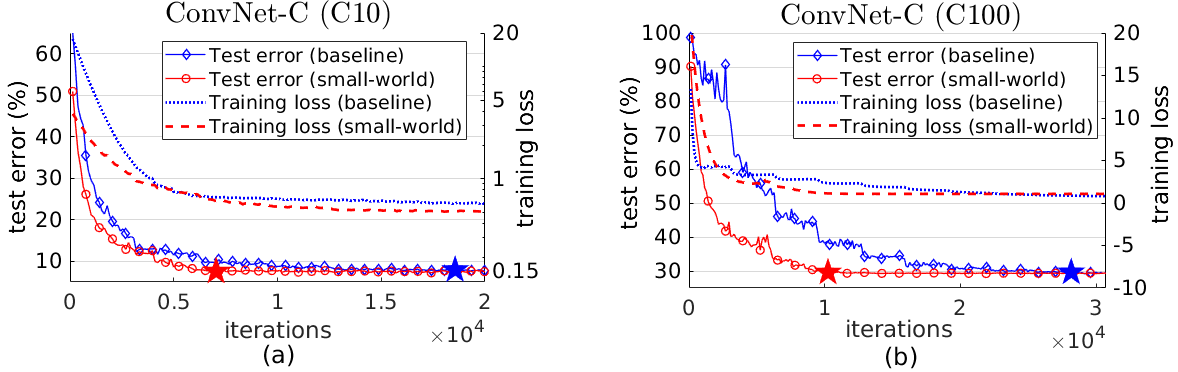}
    \vspace{-0.5em}
    \caption{Test error and training loss versus iterations for a ConvNet-C model and the rewired \sys{} trained on (a) CIFAR10 and (b) CIFAR100 datasets. The $\star$ marker denotes the point of convergence to final test accuracy for the models.}
    \label{fig:cifars}
\end{figure*}

\vspace{-0.1cm}
\subsection{Results on CIFAR}\label{sec:cifar}
\subsubsection{ConvNet-C}
\vspace{-0.1cm}
\noindent\textbf{Training.} We train the ConvNet-C~\cite{simonyan2014very} model on C10 and C100 benchmarks with a batch size of 128. To prevent overfitting, dropout layers with a rate of $0.4$ are added after $BN$ layers with no $MaxPool$, and a rate of $0.5$ before the first $FC$ layer. The small-world model is constructed using the same configuration of layers as the baseline, including the dropout layers. We use Stochastic-Gradient-Descent (SGD) optimizer with Nesterov, $0.9$ momentum, and a $5e-4$ weight decay. Models are trained for $2e+4$ and $3e+4$ iterations on C10 and C100, respectively. The initial learning rate is set to 0.01 for both datasets and learning rate is decayed by 0.5 upon optimization plateau. 



\vspace{-0.8cm}
\noindent\textbf{Convergence.} Fig.~\ref{fig:cifars}-(a) illustrates the test error and training loss of the baseline and \sys{}s as two representatives of the convergence speed. Similarly, for C100 benchmark, the corresponding convergence curve is presented in Fig.~\ref{fig:cifars}-(b). While these figures qualitatively demonstrate the effectiveness of our methodology, we provide a quantitative measure for a solid comparison between \sys{} and the baseline. We investigate several points corresponding to various test accuracies and compare the two models' convergence time to these points. Tab.~\ref{tab:cifar_accuracy} summarizes the per-accuracy speed-up of \sys{} over the baseline model. As seen, the speed-up varies for different accuracies, however, for all test accuracies, \sys{} requires a substantially fewer number of iterations for convergence. At final saturation point (marked by $\star$ on Fig.~\ref{fig:cifars}), both models achieve comparable accuracies while \sys{} enjoys a $2.64\times$ and $2.82\times$ reduction in convergence time for C10 and C100 datasets, respectively.

\vspace{-0.2cm}
\begin{table}[h]
\caption{Point-wise comparison of convergence speed-up for a \sys{} and its equivalent baseline network (ConvNet-C) on CIFAR benchmarks.}\label{tab:cifar_accuracy}
\vspace{-0.1cm}
\resizebox{1\columnwidth}{!}{
\begin{tabular}{l|l|l|c|c|c|c|c}
\hline
\multirow{5}{*}{\rotatebox[origin=c]{90}{\textbf{CIFAR10}}} & \multirow{2}{*}{\textbf{Baseline}}                & \multicolumn{1}{c|}{Test Error (\%)} & 24.21 & 17.80 & 9.22  & 8.51  & \textbf{\color{blue}7.56}  \\ \cline{3-8} 
 &  & Iterations & 1408    & 2560    & 8704    & 11008    & 18560   \\ \cline{2-8}  & \multirow{2}{*}{\textbf{\sys{}}} & \multicolumn{1}{c|}{Test Error (\%)} & 23.73 & 17.57 & 8.64  & 8.25  & \textbf{\color{blue}7.44}  \\ \cline{3-8}  &    & Iterations   & 896     & 1536    & 4992    & 5888    & 7040    \\ \hhline{~=======}   & \multicolumn{2}{l|}{\textbf{Speed-up}} & 1.57$\times$ & 1.67$\times$ & 1.74$\times$ & 1.87$\times$ & \textbf{\color{red}2.64$\times$} \\ \hline \hline
\multirow{5}{*}{\rotatebox[origin=c]{90}{\textbf{CIFAR100}}} & \multirow{2}{*}{\textbf{Baseline}}  & Test Error (\%)                      & 77.08 & 52.3  & 41.54 & 31.14 & \textbf{\color{blue}29.52} \\ \cline{3-8} &          & Iterations    & 2944    & 6144    & 9472    & 16128   & 28928   \\ \cline{2-8} & \multirow{2}{*}{\textbf{\sys{}}} & Test Error (\%)  & 76.67 & 50.57 & 40.18 & 31.15 & \textbf{\color{blue}29.26} \\ \cline{3-8} &   & Iterations  & 384     &  1408   & 3072    & 7808    & 10240    \\ \hhline{~=======}  & \multicolumn{2}{l|}{\textbf{Speed-up}} & 7.67$\times$  & 4.36$\times$  & 3.08$\times$  & 2.1$\times$   & \textbf{\color{red}2.82$\times$}  \\ \hline
\end{tabular}
}
\end{table}

\vspace{-0.4cm}
\subsubsection{DenseNet}
\vspace{-0.1cm}
DenseNets~\cite{huang2017densely} achieve state-of-the-art accuracy by connecting all neurons from different layers of a dense block with trainable (dense) parameters. Such dense connectivity pattern results in high redundancy in the parameter space and causes extra overhead on training. We show that a \sys{} with only sparse connections and much fewer parameters achieves similar results as DenseNet.

\vspace{0.1cm}
\noindent\textbf{Training.} We train a DenseNet model with 40 layers and $k=12$ (Tab.~\ref{tab:architectures}) on C10 dataset. The equivalent \sys{} is constructed by removing all long-range dense connection from the architecture and rewiring the remaining short-range edges such that each dense block transitions into a small-world structure. The \sys{} maintains the same number of layers while the inter-layer connections are implemented using sparse convolution kernels, thus incurring substantially fewer number of trainable parameters. 

\vspace{0.1cm}
We use the publicly available PyTorch implementation for DenseNets~\footnote{https://github.com/andreasveit/densenet-pytorch} and replace the model with our small-world network. Same training scheme as explained in the original DenseNet paper~\cite{huang2017densely} is used: models are trained for $19200$ iterations with a batch size of 64. Initial learning rate is $0.1$ and decays by 10 at $\frac{1}{2}$ and $\frac{3}{4}$ of the total training iterations. 

\begin{figure}[h]
    \centering
    \includegraphics[width=0.95\columnwidth]{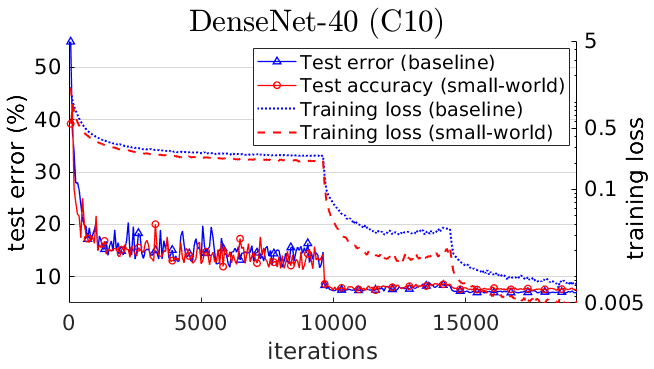}
    \vspace{-0.5em}
    \caption{Training loss and testing accuracy of the 40-layer ($k$=12)  DenseNet~\cite{huang2017densely}  with 1M parameters and our corresponding \sys{} with less than 100K parameters.}
    \label{fig:densenet}
\end{figure}

\noindent\textbf{Convergence.} Fig.~\ref{fig:densenet} demonstrates the test accuracy of the models versus the number of epochs. As can be seen, although \sys{} has much fewer parameters, both models achieve comparable validation accuracy while showing identical convergence speed. We report the computational complexity (FLOPs) of the models as the total number of multiplications performed during a forward propagation through the network. Tab.~\ref{tab:densenet_tab} compares the benchmarked DenseNet and \sys{} in terms of FLOPs and number of trainable weights in $Conv$ and $FC$ layers. We highlight that \sys{} achieves comparable test accuracy while having $10\times$ reduction in parameter space size. 

\vspace{-0.2cm}

\begin{table}[h]\centering
\vspace{-0.2cm}
\caption{Comparison of the computational complexity and model parameter space between a 40-layer DenseNet with $k$=12 and the corresponding \sys{}. In this experiment, no data augmentation is applied to either model.}\label{tab:densenet_tab}
\vspace{-0.5em}
\resizebox{0.92\columnwidth}{!}{
\begin{tabular}{l|c|c|c|c}
\hline
\textbf{Model}                   & \multicolumn{1}{l|}{\textbf{Depth}} & \multicolumn{1}{l|}{\textbf{Params}} & \multicolumn{1}{l|}{\textbf{FLOPs}} & \multicolumn{1}{l}{\textbf{Test Error}} \\ \hline
\textbf{DenseNet ($k=12$)}         & 40                         & \color{red}910K                          & 285.3M                     & 0.071                                 \\ \hline
\textbf{\sys{}} & 40                         & \color{red}98K                         & 85.5M  & 0.074   \\ \hline                     
\end{tabular}}
\vspace{-0.2cm}
\end{table}

%% file: 7_2_imagenet.tex
\subsection{Results on ImageNet}\label{sec:imagenet}
\subsubsection{AlexNet}
\vspace{-0.2cm}
\noindent\textbf{Training.} We train the AlexNet~\cite{krizhevsky2012imagenet} model on ImageNet dataset and follow the architecture provided in Caffe model zoo~\cite{caffe_alexnet} (See Tab.~\ref{tab:architectures}). In order to mitigate overfitting, we add dropout layers with probability 0.5 after each $FC$ layer (except the last). Loss minimization is performed by means of SGD with Nesterov~\cite{nesterov2007gradient} and a 0.9 momentum. We set the batch size to 64 for both models and incorporate an exponential decay for the learning rate: initial learning rate is set to $2.5e{-3}$ and the decay factor is $0.99999875$~\cite{smith2017cyclical}. 

\begin{figure}[h]
    \centering
    \includegraphics[width=0.87\columnwidth]{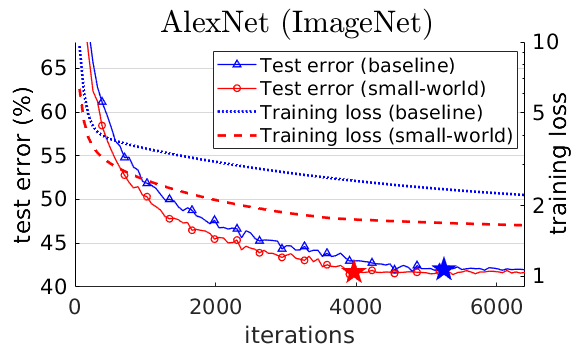}
    \vspace{-0.2cm}
    \caption{Convergence plots of an AlexNet architecture and its \sys{} on the ImageNet dataset. The $\star$ marker indicates convergence to the final test error rate.}
    \label{fig:imagenet}
\end{figure}

\noindent\textbf{Convergence.} Fig.~\ref{fig:imagenet} demonstrates the test error and training loss of the baseline and \sys{}s. As can be seen, for all values of test error, the convergence of our small-world architecture is faster. Similar to CIFAR benchmarks, to fully examine the performance of our model, we report the speed-up of \sys{} over the baseline for several values of test error. Tab.~\ref{tab:imagenet_accuracy} encloses the point-wise comparison between the benchmarked models. As indicated by our evaluations, \sys{} converges to the final test accuracy after 3776 iterations while the baseline model needs 5120  iterations, resulting in a $1.36\times$ overall speed-up. 

\begin{table}[h]
\vspace{-0.5cm}
\caption{Performance of a baseline AlexNet and its \sys{} benchmarked on ImageNet dataset. }\label{tab:imagenet_accuracy}
\vspace{-0.2cm}
\resizebox{1\columnwidth}{!}{
\begin{tabular}{l|l|l|c|c|c|c|c}
\hline
\multirow{5}{*}{\rotatebox[origin=c]{90}{\textbf{AlexNet}}} & \multirow{2}{*}{\textbf{Baseline}}                & \multicolumn{1}{c|}{Test Error (\%)} & 51.72 & 46.29 & 44.21 & 42.31 & \textbf{\color{blue}42.01} \\ \cline{3-8} &                                          & Iterations  & 1088   & 2304  & 3264  & 4416  & 5120  \\ \cline{2-8} & \multirow{2}{*}{\textbf{\sys{}}} & \multicolumn{1}{c|}{Test Error (\%)} & 51.97 & 46.49 & 44.25 & 42.31 & \textbf{\color{blue}41.55} \\ \cline{3-8} &   & Iterations                           & 768   & 1664  & 2368  & 3520  & 3776  \\ \hhline{~=======}
      & \multicolumn{2}{l|}{\textbf{Speed-up}}                                                   & 1.42$\times$  & 1.38$\times$  & 1.38$\times$  & 1.25$\times$  & \textbf{\color{red}1.36$\times$}  \\ \hline
\end{tabular}}
\end{table}

\vspace{-0.5cm}
\subsubsection{ResNet}
\vspace{-0.1cm}
\noindent\textbf{Training.} We adopt the training scheme in the original ResNet paper~\cite{he2016deep}. To build the \sys{}, we first remove all shortcut and bottleneck connections from the model. We then rewire the connections in the acquired plain network such that it becomes small-world. No dropout is used for the baseline and \sys{}s. Batch size is set to 128 and we use SGD with $0.9$ momentum and weight decay of $1e-4$. Initial learning rate is set to 0.1 and decays by $0.1$ when the accuracy plateaus. We train the models for $9e+5$ iterations and report single-crop accuracies. 

\begin{figure}[h]
    \centering
    \includegraphics[width=0.9\columnwidth]{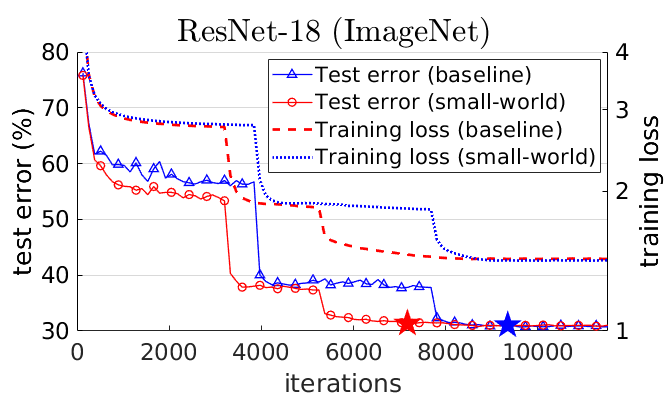}
    \vspace{-0.2cm}
    \caption{Test error and training loss across training iterations for ResNet-18 on ImageNet dataset. Convergence to minimum error rate is shown with a $\star$ marker.}
    \label{fig:resnet}
\end{figure}

\noindent\textbf{Convergence.} Test error and training loss for baseline ResNet and the \sys{} are shown in Fig.~\ref{fig:resnet}. As seen, \sys{} achieves both higher accuracy and higher convergence speed throughout training. For a more quantitative comparison, we enclose point-wise speed-ups for various iterations and test errors in Tab.~\ref{tab:resnet_accuracy}. As evident from the results, systematic restructuring of long edges in \sys{} allows for a better convergence speed compared to the replicated blocks in the baseline ResNet.

\begin{table}[h]
\centering
\vspace{-0.5cm}
\caption{Point-wise convergence comparison of a baseline ResNet-18 and its \sys{} equivalent on ImageNet dataset. }\label{tab:resnet_accuracy}
\vspace{-0.1cm}
\resizebox{0.93\columnwidth}{!}{
\begin{tabular}{l|c|l|c|c|c|c}
\hline
\multirow{5}{*}{\rotatebox[origin=c]{90}{\textbf{ResNet-18}}} & \multirow{2}{*}{\textbf{BaseLine}}                & Test Error (\%) & 60.37 & 56.94 & 37.91 & \textbf{\color{blue}31.72} \\ \cline{3-7} &  & Iterations & 1792  & 3456  & 7424  & 9344  \\ \cline{2-7}  & \multirow{2}{*}{\textbf{\sys{}}} & Test Error (\%) & 59.63 & 56.76 & 37.86 & \textbf{\color{blue}31.68} \\ \cline{3-7} &  & Iterations      & 512   & 768   & 3584  & 7168  \\ \hhline{~======}  & \multicolumn{2}{l|}{\textbf{Speed-up}}              & 3.50$\times$ & 4.50$\times$ & 2.07$\times$ & \textbf{\color{red}1.31}$\times$ \\ \hline
\end{tabular}}
\end{table}

%% file: 8_discussion.tex
\vspace{-0.3cm}
\section{Discussion on Long-range Connections}\label{sec:discussion}
\begin{figure}[h]
    \centering
    \includegraphics[width=1\columnwidth]{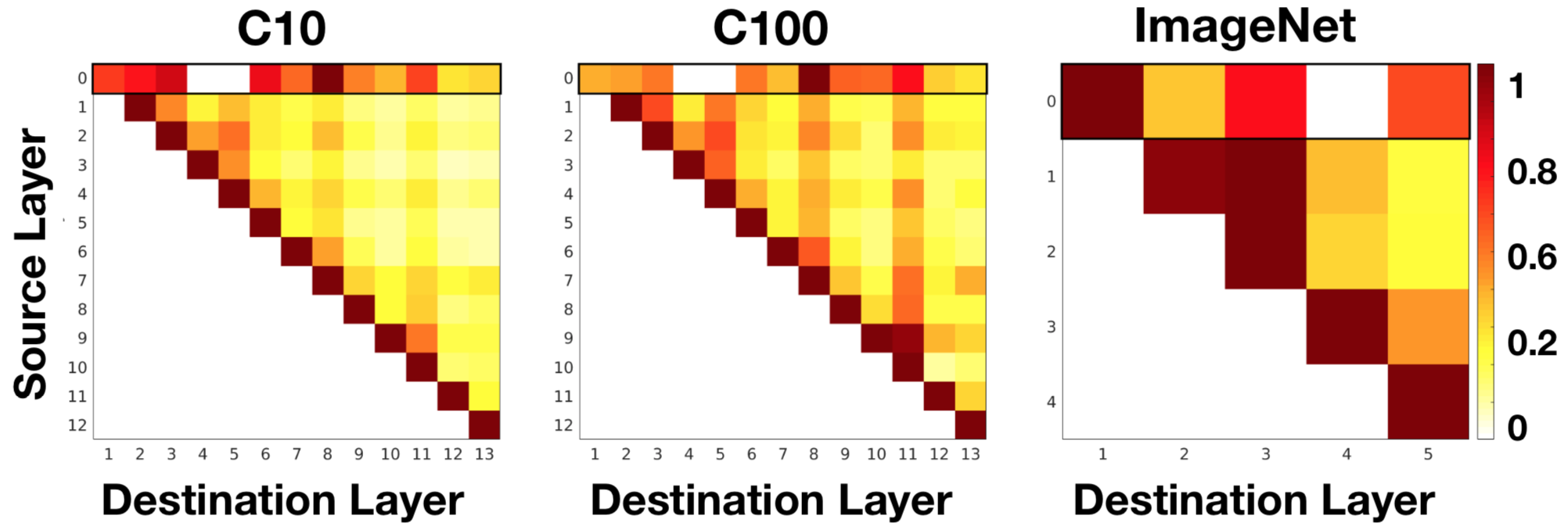}
    \caption{Visualization of average absolute value of trained weights within $Conv$ layers of a \sys{}. Colors encode the connectivity strength between layers with red being the strongest and white denoting no connection. The marked rows with black box borders correspond to the input layer of the networks.}
    \label{fig:heatmap}
\end{figure}

\vspace{-0.1cm}
The selected small-world structure for a given CNN has two main characteristics, namely high clustering of nodes and small average path length between neurons across layers. We postulate that such qualities render the SWN desirable during training due to the enhanced information flow paths existent in these efficiently-connected networks. To examine our hypothesis, we visualize the weights connecting different layers of the trained \sys{} for C10, C100 (ConvNet-C), and ImageNet (AlexNet) benchmarks. Fig.~\ref{fig:heatmap} presents a heat map of the average absolute values of weights connecting each pair of $Conv$ layers.


Each square at position $(l_1, l_2)$ of the heatmap represents the strength of the connections between layers $l_1$ and $l_2$ where $l_0$ denotes network input. Color shades of orange, red and maroon indicate strong inter-layer dependency while the white color indicates that no connections are present between the corresponding layers in \sys{}. We summarize our observations based on the heat map as the following:
\vspace{-0.5em}
\begin{enumerate}
    \item Each layer has strong connections to its non-subsequent layers indicating that long-range edges established in \sys{} are crucial to performance.\vspace{-0.5em}
    \item The input layer has spread weights across all layers of the network which demonstrates the importance of connections between earlier and deeper layers.\vspace{-0.5em}
    \item \sys{} preserves the strong connections between one layer and the immediately proceeding layer, thus, maintaining the conventional CNN data flow.
\end{enumerate}


%% file: 9_conclusion.tex
\section{Conclusion}
We propose a novel methodology that adaptively modifies conventional feed-forward DL models to new architectures, called \sys{}, that fall into the category of small-world networks\textemdash a class of complex graphs used to study real-world models such as human brain and the neural networks of animals. By leveraging the intriguing features of small-world networks, e.g., enhanced signal propagation speed and synchronizability, \sys{}s enjoy enhanced data flow within the network, resulting in substantially faster convergence speed during training. Our small-world models are implemented via sparse connections from each layer in the traditional CNN to all the succeeding layers. Such sparse convolutions enable \sys{}s to benefit from long-range connections while mitigating the redundancy in the parameter space existent in prior art. As our experiments demonstrate, \sys{}s are able to achieve state-of-the-art accuracy in $\approx 2.1\times$ lower number of training iterations, on average. Furthermore, compared to a densely-connected architecture, \sys{}s achieve comparable accuracy while having $10\times$ reduction in the number of parameters. In summary, due to their optimal graph connectivity and fast response to training, \sys{}s can be advantageous for smart vision applications.